\documentclass[runningheads]{llncs}
\usepackage{booktabs}       
\usepackage{amsfonts}       
\usepackage{multirow}
\usepackage{listofitems}
\usepackage{graphicx}
\usepackage[caption=false]{subfig}
\usepackage{pbox}

\usepackage{comment}
\usepackage{amsmath,amssymb} 
\usepackage{color}

\newcommand{\etal}{\textit{et al.}}

\begin{document}
\pagestyle{headings}
\mainmatter
\def\ECCVSubNumber{6267}  

\title{Hierarchical Opacity Propagation for Image Matting} 


\titlerunning{HOP Matting}
%
\author{Yaoyi Li \and Qingyao Xu\and Hongtao Lu}
\authorrunning{Y. Li et al.}
%
\institute{Department of Computer Science and Engineering,\\ Shanghai Jiao Tong University, China
}
\maketitle

\begin{abstract}
Natural image matting is a fundamental problem in computational photography and computer vision. Deep neural networks have seen the surge of successful methods in natural image matting in recent years. In contrast to traditional propagation-based matting methods, some top-tier deep image matting approaches tend to perform propagation in the neural network implicitly. A novel structure for more direct alpha matte propagation between pixels is in demand. To this end, this paper  presents a hierarchical opacity propagation (HOP) matting method, where the opacity information is propagated in the neighborhood of each point at different semantic levels. The hierarchical structure is based on one global and multiple local propagation blocks. With the HOP structure, every feature point pair in high-resolution feature maps will be connected based on the appearance of input image. We further propose a scale-insensitive positional encoding tailored for image matting to deal with the unfixed size of input image and introduce the random interpolation augmentation into image matting. Extensive experiments and ablation study show that HOP matting is capable of outperforming state-of-the-art matting methods.

\end{abstract}

\section{Introduction}
Natural image matting is the problem of separating the foreground object from background. The digital image matting treats the input image as a composition and aims to estimate the opacity of foreground image. The predicted results are alpha mattes which indicate the transition between foreground and background at each pixel \cite{wang2008image}.

\begin{figure}[t]
	\centering	 
	\subfloat[Image]{
		\centering
		\includegraphics[width=.24\columnwidth]{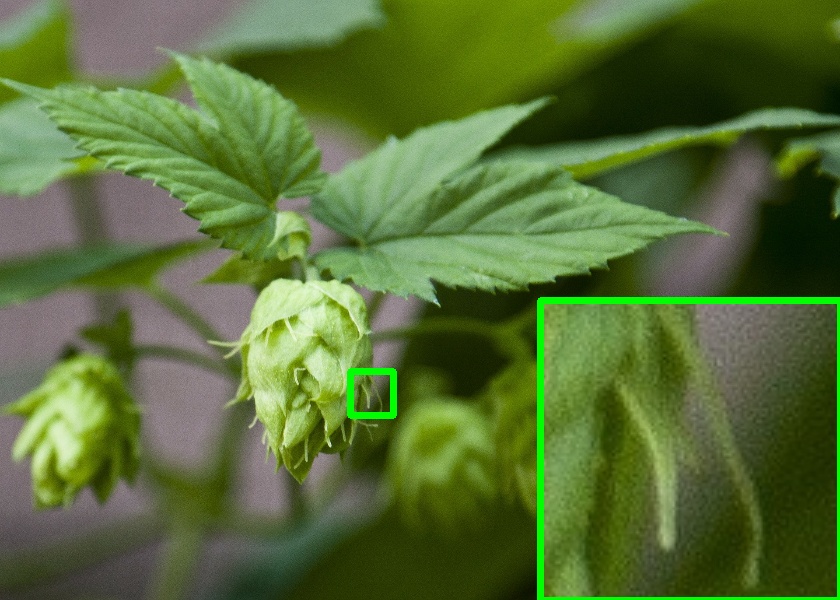}
	}
	\centering	
	\subfloat[Trimap]{
		\centering
		\includegraphics[width=.24\columnwidth]{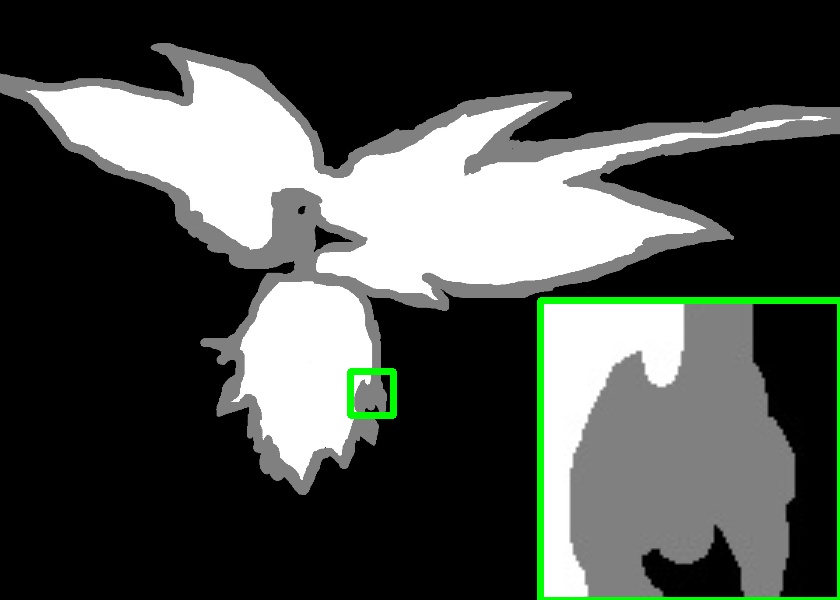}
	}
	\centering	
	\subfloat[Deep Matting]{
		\centering
		\includegraphics[width=.24\columnwidth]{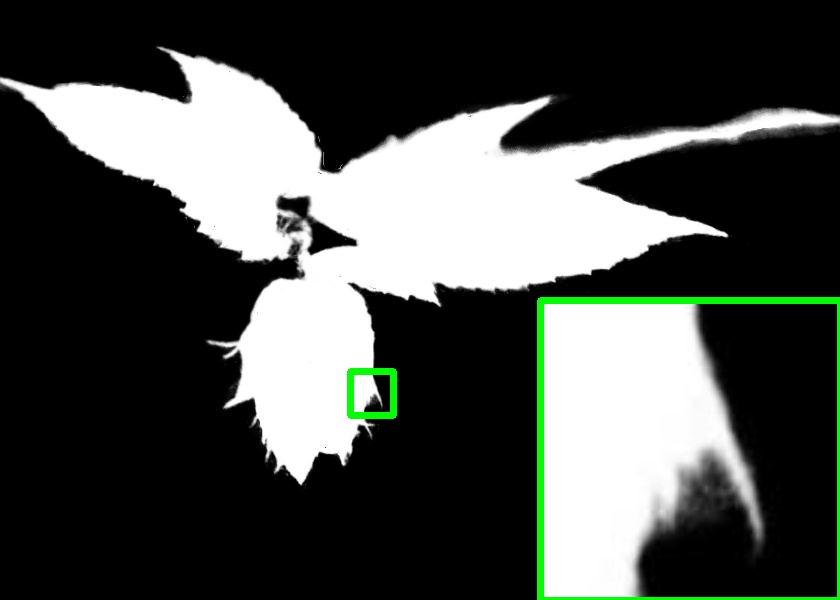}
	}

	\centering	
	\subfloat[IndexNet Matting]{
		\centering
		\includegraphics[width=.24\columnwidth]{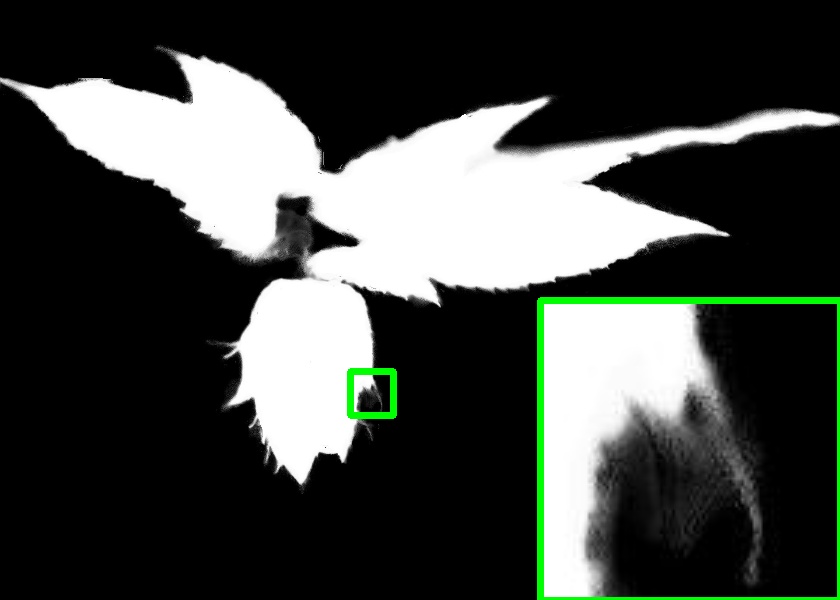}
	}
	\centering	
	\subfloat[Context-aware]{
		\centering
		\includegraphics[width=.24\columnwidth]{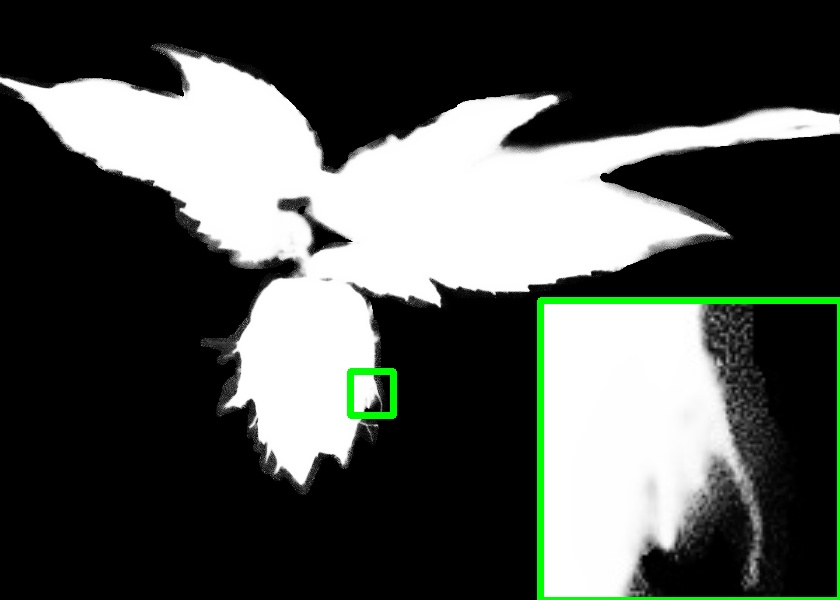}
	}
	\centering	
	\subfloat[Ours]{
		\centering
		\includegraphics[width=.24\columnwidth]{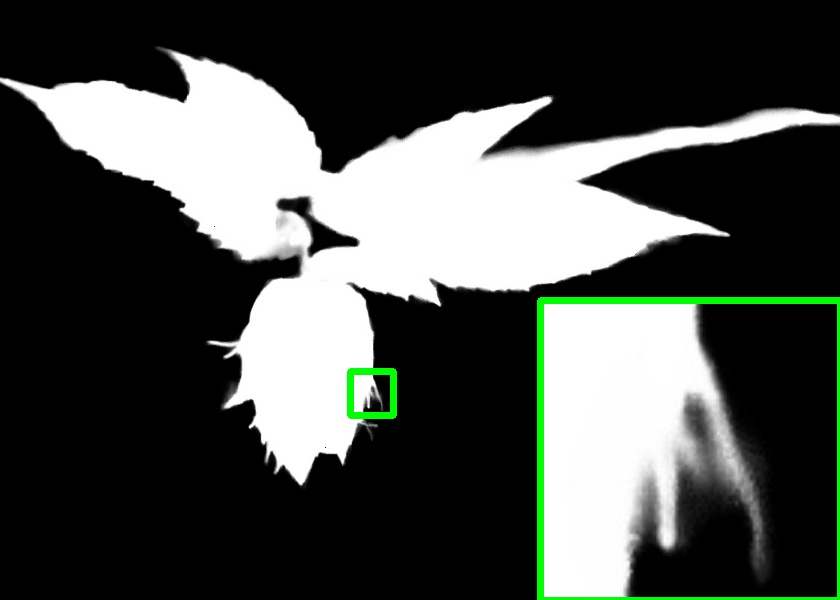}
	}
	\caption{The demonstration on a real world hops image\protect\footnotemark.}
	\label{fig:demo}
\end{figure}

Formally, the observed RGB image $I$ is modeled as a convex combination of the foreground image $F$ and background $B$ \cite{chuang2001bayesian,wang2008image}:

\begin{equation}
I_p = \alpha_pF_p + (1-\alpha_p)B_p, \quad \alpha_p \in [0,1],
\label{eq:matting}
\end{equation}
where $\alpha_p$ denotes the alpha matte value to be estimated at position $p$. The original definition of image matting is an ill-defined problem. Therefore, in most matting tasks, trimap images (Figure \ref{fig:demo}(b)) are provided as a coarse annotation, indicating the known foreground and background region as well as the unknown region to be predicted.

Typically, conventional propagation-based matting algorithms \cite{levin2008closed,he2010fast,lee2011nonlocal,chen2013knn,aksoy2017designing} generate the alpha matte by transmitting opacity or transparency between pixels by referring to the appearance similarity in the input image.
This inductive bias is leveraged by some deep learning based matting approaches implicitly to benefit their matting results \cite{samplenet,cai2019disentangled}.
SampleNet \cite{samplenet} performs propagation by the inpainting network \cite{yu2018generative} adopted in their method for foreground and background estimation instead of opacity prediction. Moreover, the propagation of inpainting part is only carried out on the semantically strong features rather than high-resolution features.  In AdaMatting \cite{cai2019disentangled}, the authors introduced convolutional long short term memory (LSTM) networks \cite{xingjian2015convolutional} into their network as the last stage for propagation, in which the propagation is performed based on the convolution and memory kept in the cell of ConvLSTM.
The difference between ConvLSTM and direct propagation is analogous to the distinction between LSTM \cite{hochreiter1997long} and transformer \cite{vaswani2017attention}.

\footnotetext{Photograph by Lisa Baird from Pixabay.}

In this paper, we present a novel hierarchical opacity propagation structure in which the opacity messages are transmitted across different semantic feature levels. The proposed structure is constructed by two propagation blocks, namely global HOP block and local HOP block. The HOP blocks perform information transmission by the mechanism of attention and aggregation \cite{bahdanau2014neural} as transformer does \cite{vaswani2017attention}. Notwithstanding, both global and local HOP blocks are designed as two-source transformers. More concretely, The relation between nodes in attention graph is computed from appearance feature and the information to be propagated is the opacity feature, which contrasts with self-attention \cite{vaswani2017attention,ramachandran2019stand} or conventional attention \cite{bahdanau2014neural,xu2015show}.
With the help of HOP structure, our network learns to predict the opacity over context on the low-resolution but semantically strong features by global HOP block and refine blurring artifacts on the high-resolution features by local HOP block. A demonstration of our method as well as three state-of-the-art approaches on a real world image is shown in Figure \ref{fig:demo}.
In addition to the HOP structure, we present a scale-insensitive positional encoding to handle the variant size of input images based on the relative positional encoding \cite{dai2019transformer,ramachandran2019stand}. The random interpolation is also introduced into our method as an augmentation for matting to further boost the performance.

Specifically, our proposed method differ from previous deep image matting approaches in following aspects:

1. We propose a novel hierarchical propagation architecture for image matting based on a series of global and local HOP blocks, leveraging opacity and appearance information in different semantic levels for information propagation.

2. We introduce the random interpolation augmentation into the training of deep image matting and empirical evaluation results show that the augmentation makes a remarkable gain in performance.

3. Experiments on both Composition-1k testing set and alphamatting.com dataset demonstrate that the proposed approach is competitive with the other state-of-the-art  methods.

\section{Related Work}
Most of the natural image matting approaches can be  broadly categorized as propagation-based methods \cite{levin2008closed,he2010fast,lee2011nonlocal,chen2013knn,aksoy2017designing}, sampling-based methods \cite{wang2007optimized,gastal2010shared,he2011global,feng2016cluster} and learning-based methods \cite{xu2017deep,lutz2018alphagan,lu2019indices,samplenet,hou2019context,cai2019disentangled}. In this section, we will review some deep learning based methods that are highly related to our work.


\begin{figure}[t]
	\centering	 
	\subfloat[]{
		\centering
		\includegraphics[width=0.6\linewidth]{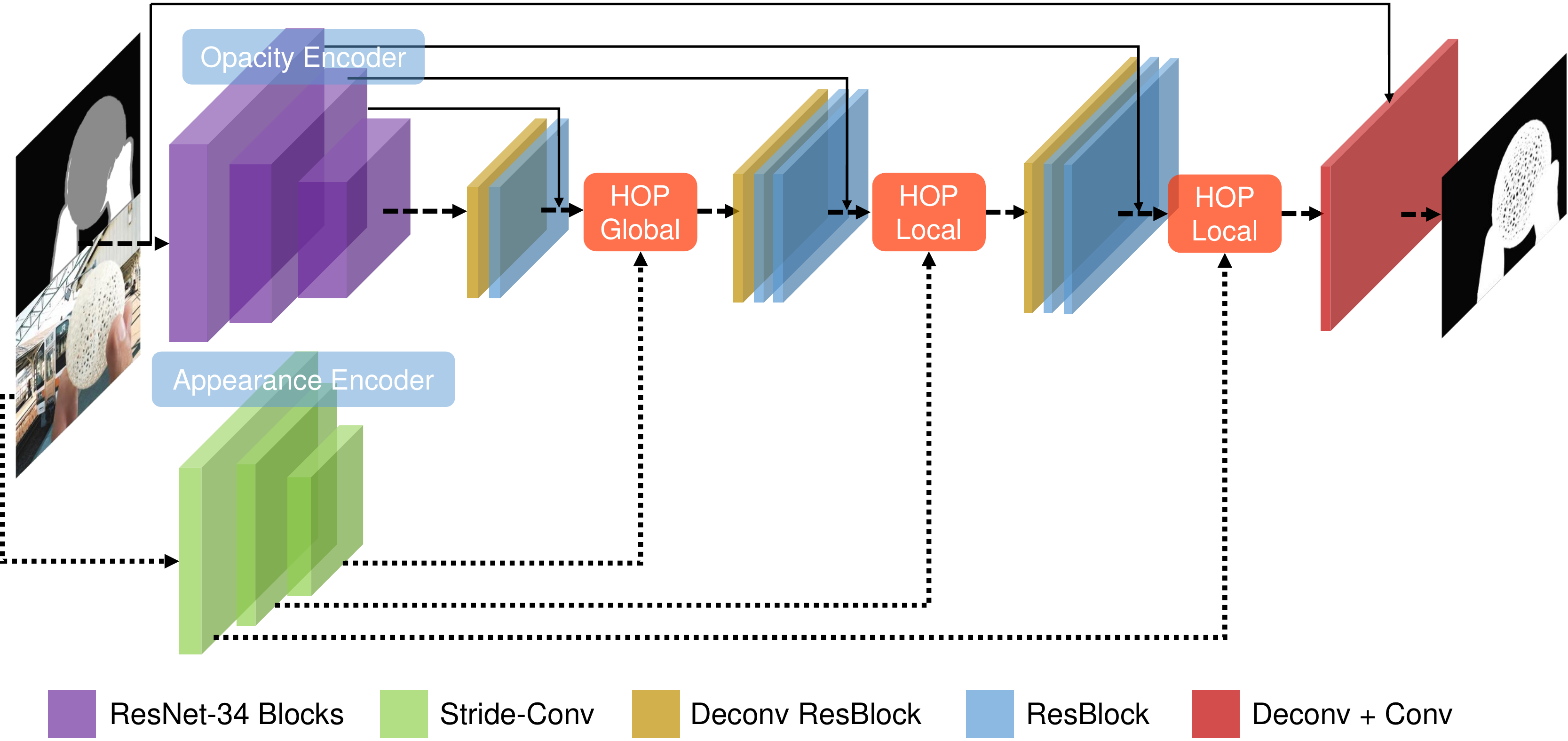}
	}
	\centering	 
	\subfloat[]{
		\centering
		\includegraphics[width=0.3\linewidth]{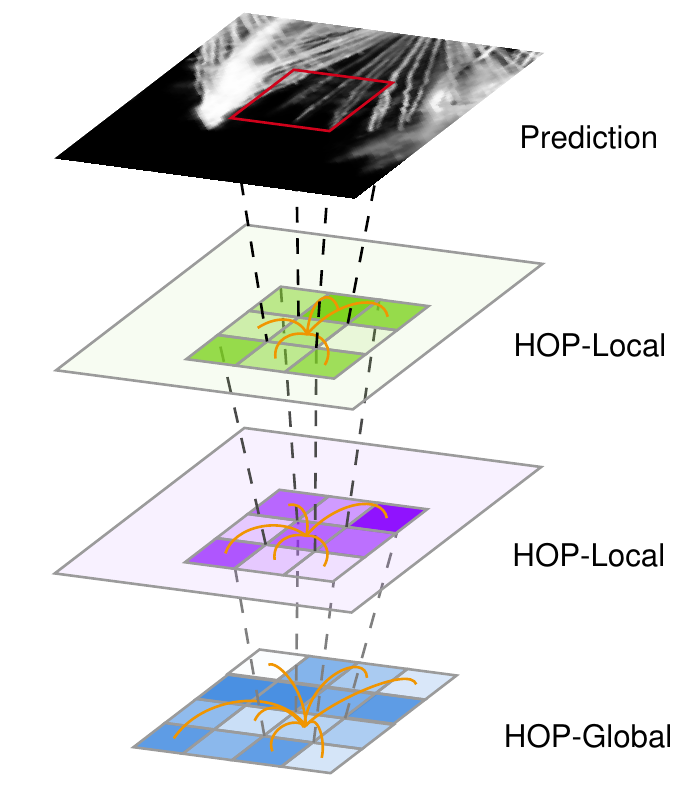}
	}
	\caption{Diagrams of our proposed HOP structure. For the ease of representation, only a 4-scale-level decoder and 2 local HOP blocks are shown. In our implementation, we have a 5-level decoder and 3 local HOP blocks. (a) The architecture of our network. The appearance encoder branch only takes RGB image as the input. (b) The schematic diagram of hierarchical opacity propagation on feature maps from different semantic level. Orange lines indicate the feature propagation.}
	\label{fig:diag}
\end{figure}

General deep learning based methods utilize deep network to directly predict the alpha mattes with the given images and trimaps.
Cho \etal \cite{cho2019deep} used an end-to-end CNN which toke mattes from closed-form matting \cite{levin2008closed} and KNN matting \cite{chen2013knn} together with normalized RGB images as the input to better utilize local and nonlocal structure information.
Deep Matting is proposed in \cite{xu2017deep} as  a two-stage network to predict the alpha matte. In the first stage they fed their deep convolutional encoder-decoder network with images and corresponding trimaps. In the second stage, the coarse predicted alpha matte was refined by a shallow network.
Lutz \etal \cite{lutz2018alphagan} introduced GAN into image matting problem and used dilated convolution to capture the global context information.
Tang \etal \cite{tang2019very} proposed a network called VDRN involving a deep residual encoder and a sophisticated decoder.
AdaMatting \cite{cai2019disentangled} divided the matting problem into two tasks, trimap adaptation and alpha estimation, and used a deep CNN with two distinct decoder branches to handle the tasks in a multitasking manner.
IndexNet Matting \cite{lu2019indices} considered indices in pooling as a function of feature map and proposed an index-guided encoder-decoder network using the index pooling and upsampling guided by learned indices.
Hou and Liu \cite{hou2019context} proposed a context-aware network to predict both foreground and alpha matte. Context-aware Matting employed two encoders, one for local feature and the other one for context information. GCA Matting \cite{li2020natural} introduced guided contextual attention mechanism to analogize image inpainting processing in matting.

Instead of predicting alpha matte directly, some methods show that changing the degrees of freedom will influence the performance of their network.
Tang \etal \cite{samplenet} proposed SampleNet to estimate foreground, background and alpha matte step by step, rather than predict them simultaneously.
Zhang \etal \cite{Zhang2019ALF} used two decoder branches to estimate foreground and background respectively, then they used a fusion branch to obtain the result.

All methods mentioned above relies heavily on the trimap as a input to reduce the resolution space. However recently, for some specific practical problems, methods that leverage the image semantic information and work well without any trimap proposed.
Shen \etal \cite{shen2016deep} combined an end-to-end CNN with closed-form matting \cite{levin2008closed} to automatically generate the trimap of portrait image and then obtained desired alpha mattes.
Chen \etal \cite{chen2018semantic} proposed a Semantic Human Matting that integrated a semantic segmentation network with a matting network to automatically extract the alpha matte of humans.


\section{Method}
To achieve the direct opacity information propagation all over the input image at a high resolution level, we propose the novel structure of  hierarchical opacity propagation, in which the neural network can be regarded as a multi-layer graph convolutional network \cite{kipf2016semi} with different graphs and the opacity can be propagated between every two pixels. 
In this section, we will introduce our hierarchical opacity propagation blocks first, and then propose the scale-insensitive positional encoding for our HOP block. Afterwards, the implementation details will be described.

\subsection{Hierarchical Opacity Propagation Block}
Typically, an non-local block \cite{wang2018non} or a transformer \cite{vaswani2017attention} is capable of carrying out the information propagation globally by its self-attention mechanism \cite{lin2017structured}. However, there are two flaws if we adopt the original non-local block or image transformer in a natural image matting method directly. On one hand, the non-local block is computationally expensive. Although some modification were proposed to reduce the computation and memory consumption \cite{zhu2019asymmetric}, it is still infeasible to propagate the opacity information on a high-resolution  feature map, which is required by image matting. On the other hand, both non-local and transformer build a complete graph  on the input feature map and the edge weights are generated from the input feature of nodes. In image matting tasks, the feature of each node is the opacity information. It is straightforward to propagate  the semantic feature based on relationship between different feature nodes in some semantic tasks like video classification \cite{wang2018non}, semantic segmentation \cite{zhu2019asymmetric} or image inpainting \cite{yu2018generative}, whereas the propagation in natural image matting requires more non-semantic appearance information than the semantic features.

\begin{figure}[t]
	\centering	 
	\pbox[t]{.24\textwidth}{
		\includegraphics[height=.35\textwidth]{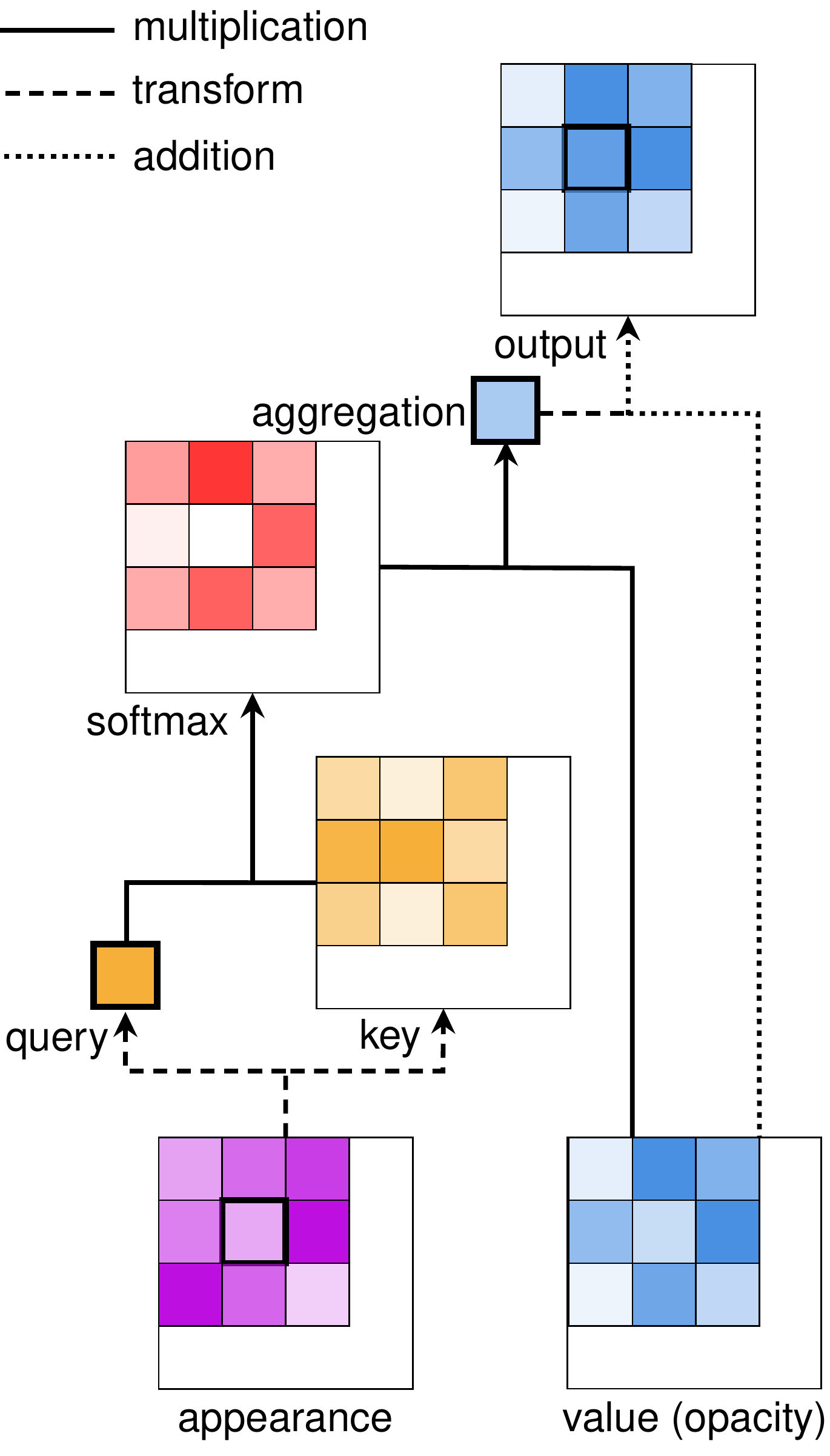}\\
		\centering (a) local HOP block
	}
	\pbox[t]{.24\textwidth}{
		\includegraphics[height=.35\textwidth]{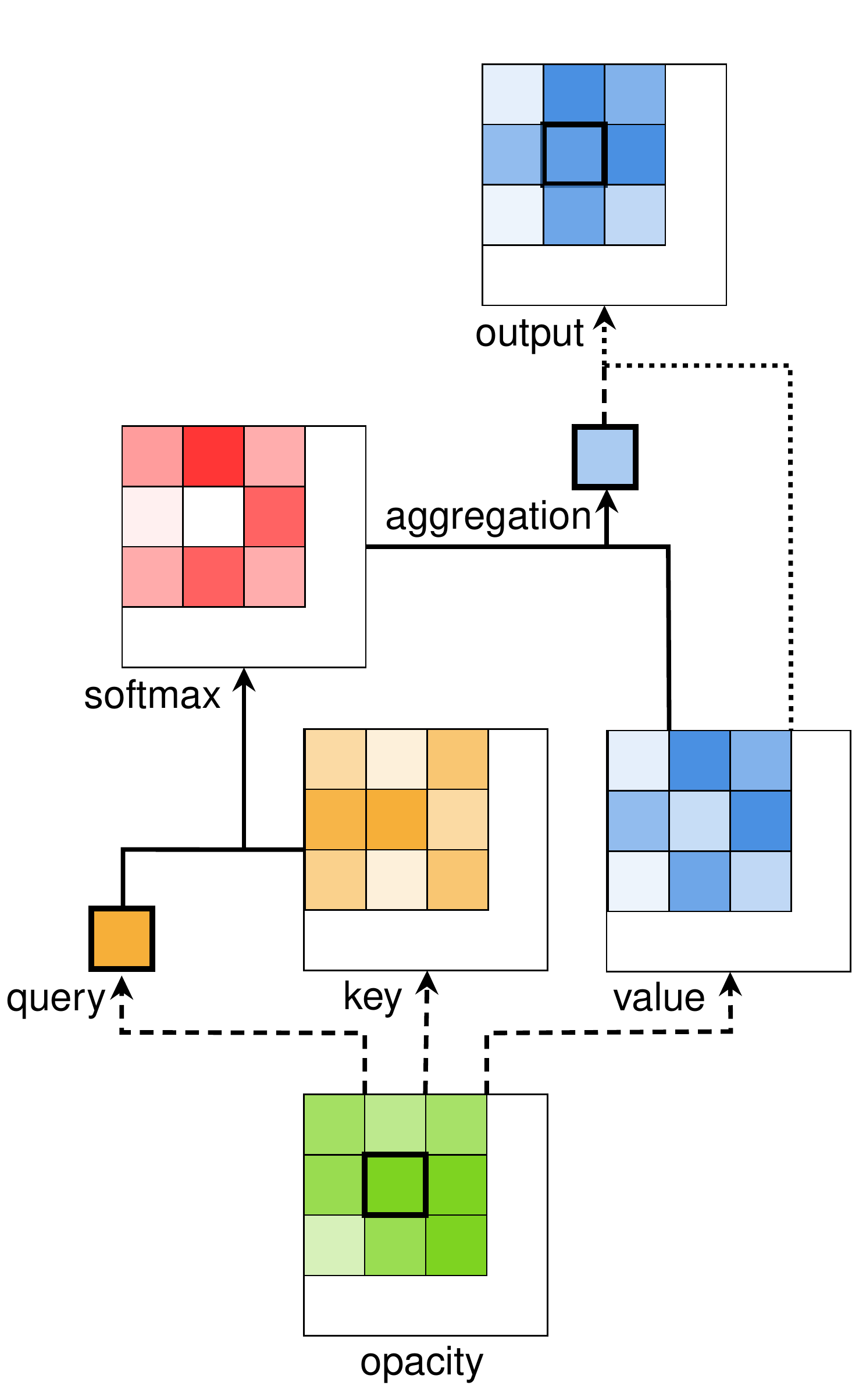}\\
		\centering (b) local self-attention
	}
	\pbox[t]{.24\textwidth}{
		\includegraphics[height=.35\textwidth]{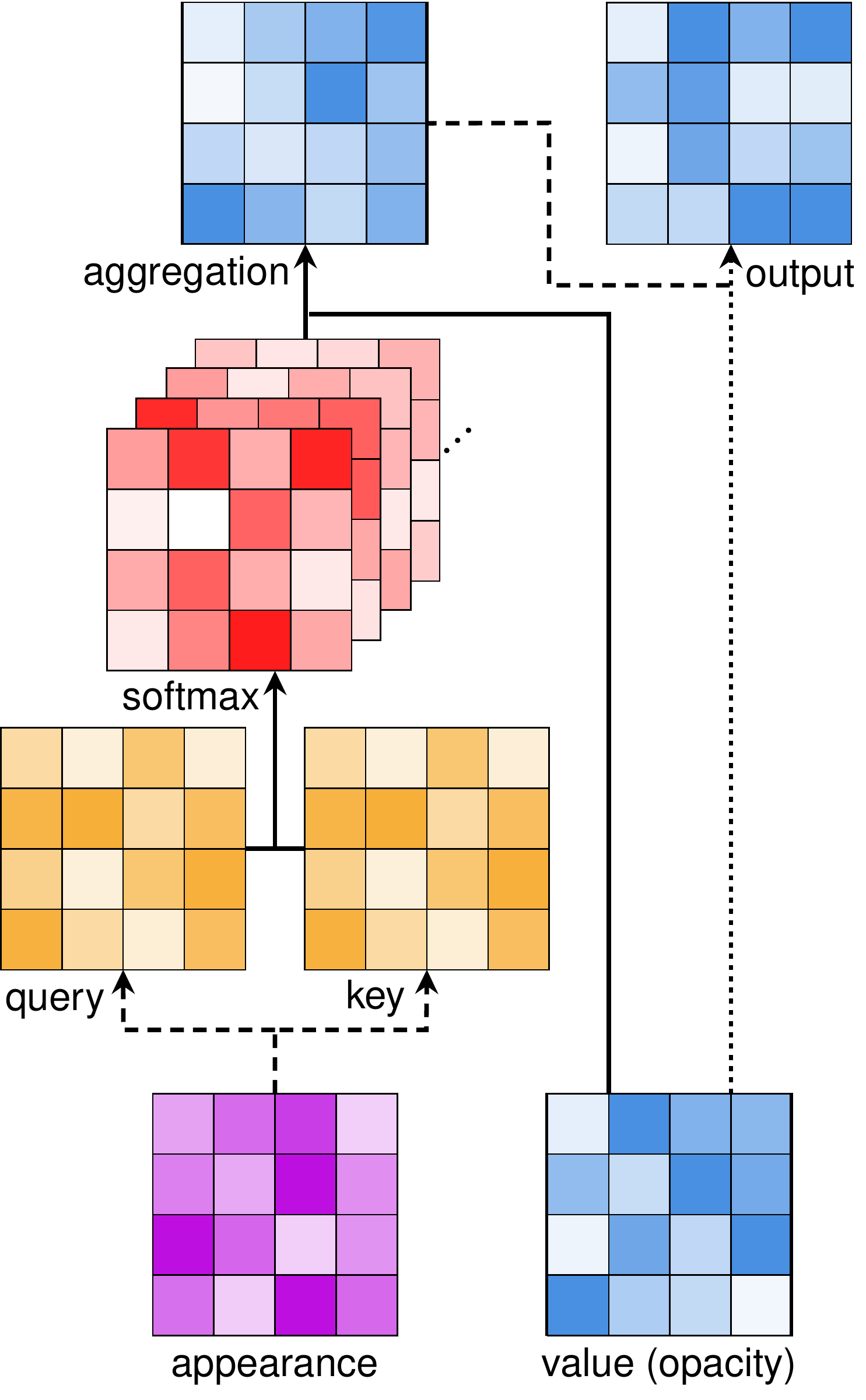}\\
		\centering (c) global HOP block
	}
	\pbox[t]{.24\textwidth}{
		\includegraphics[height=.35\textwidth]{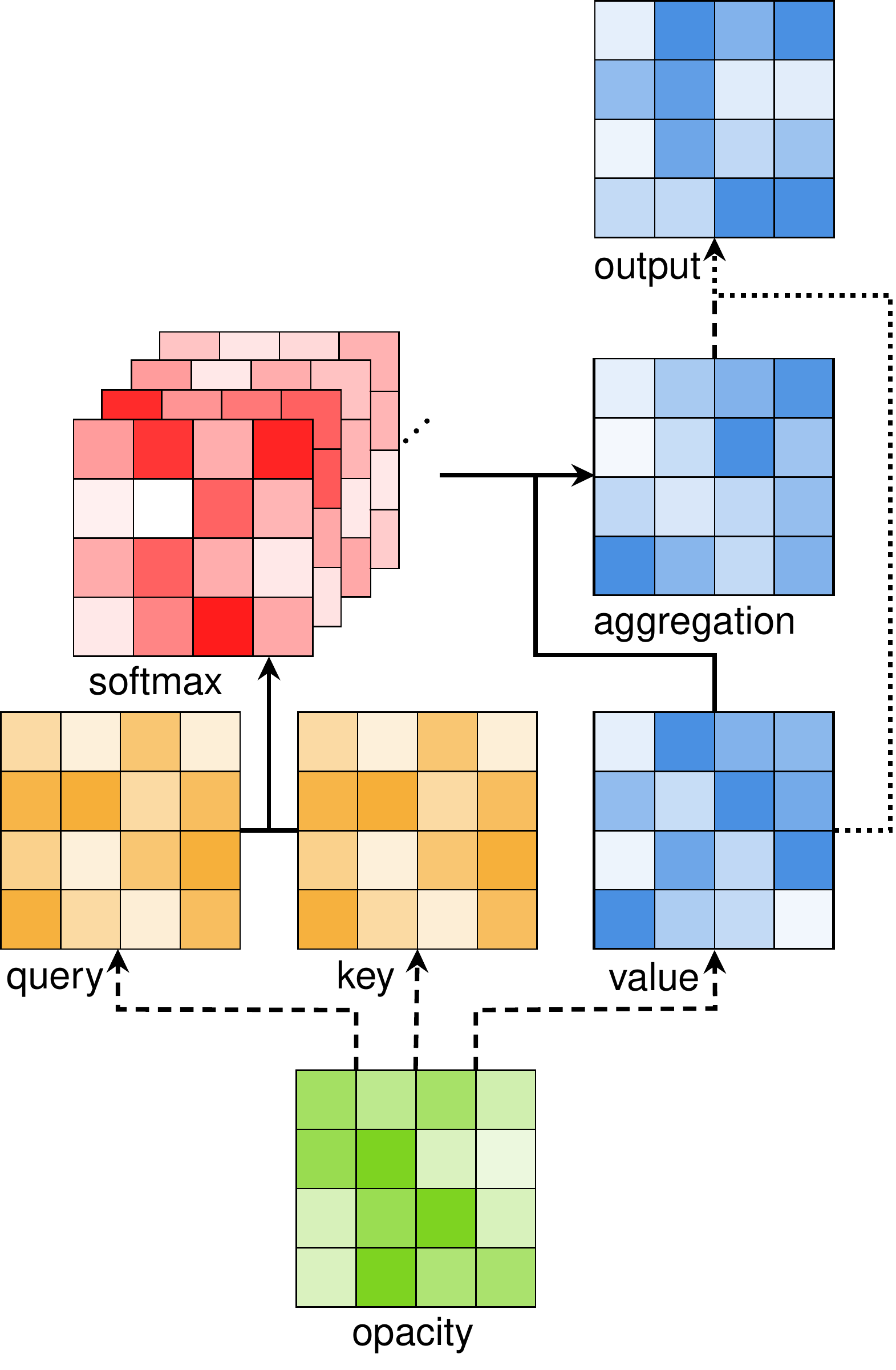}\\
		\centering (d) global self-attention
	}
	\caption{The diagrams of detailed structure of local HOP block, local self-attention block, global HOP block and global self-attention block. For these two local blocks, we only show one query data point in the diagrams for the ease of presentation.}
	\label{fig:block}
\end{figure}

Inspired by the current success of sparse transformer \cite{child2019generating,huang2019interlaced}, local self-attention \cite{ramachandran2019stand} and guided contextual attention \cite{li2020natural}, our proposed hierarchical opacity propagation structure involves two different propagation blocks, namely global HOP block and local HOP block, in which the appearance and opacity prediction are leveraged together in the propagation.

The network architecture of our proposed HOP is shown in Figure \ref{fig:diag}(a). In our method, there are two encoder branches, one for opacity information source and the other one for image appearance source. Assuming the feature maps from opacity encoder and appearance encoder are denoted by $F^O\in\mathbb{R}^{HW\times C}$ and $F^A\in\mathbb{R}^{HW\times C}$ respectively, and feature points at position $(i,j)$ are $f^O_{(i,j)}\in\mathbb{R}^{C}$ and $f^A_{(i,j)}\in\mathbb{R}^{C}$, the global HOP block can be defined as follows:
\begin{equation}
\begin{aligned}
q_{(i,j)} =& W_{QK}f^A_{(i,j)},\\ 
k_{(x,y)} =& W_{QK}f^A_{(x,y)}, \\
a_{(i,j),(x,y)} =& \mathop{\mathrm{softmax}}_{(x,y)}(\frac{q_{(i,j)}^Tk_{(x,y)}}{\|q_{(i,j)}\|\|k_{(x,y)}\|}),\\
g_{(i,j)} =& W_{out}(\sum_{(x,y)}a_{(i,j),(x,y)}f^O_{(x,y)}) + f^O_{(i,j)},
\end{aligned}
\end{equation}
where the $W_{QK}$ is the linear transformation for both key and query, $W_{out}$ is the transformation to align the propagated information with input feature map $F^O$, and the softmax is operated along $(x,y)$ dimension. Additionally, $F^O$ is the value term in this attention mechanism without any transformation, and we can also regard $v_{(x,y)}=W_{out}f^O_{(x,y)}$ as the value term. Figure \ref{fig:block}(c) demonstrates the detailed structure of global HOP block. 

In global HOP block, the key and value, as distinct from the self-attention \cite{lin2017structured,vaswani2017attention} or conventional attention mechanism \cite{bahdanau2014neural,xu2015show}, are homoplasy. In self-attention mechanism, all of the query, key and value are computed from the same feature, and in conventional attention mechanism the key and value are from the same place. However, in HOP blocks, the query and key share the same original feature of appearance source and the value item has a distinct source from opacity feature.

Similarly, we can formulate the local HOP block which only attends to the local neighborhood of each feature point:
\begin{equation}
\begin{aligned}
a_{(i,j),(x,y)} = &\mathop{\mathrm{softmax}}_{(x,y)\in\mathcal{N}((i,j), s)}(\frac{q_{(i,j)}^Tk_{(x,y)}}{\|q_{(i,j)}\|\|k_{(x,y)}\|}),\\
g_{(i,j)} = &W_{out}(\sum_{(x,y)\in\mathcal{N}((i,j), s)}a_{(i,j),(x,y)}f^O_{(x,y)})\\& + f^O_{(i,j)},
\end{aligned}
\end{equation}
where $\mathcal{N}((i,j), s)$ is the neighborhood of position $(i,j)$ with a window size of $s$.

In our propagation graph, each node has two different features, opacity and appearance. The appearance feature is only utilized to generate the edge weight of the graph and opacity feature is the de facto information to be propagated. The difference between HOP block and self-attention can also be seen from Figure \ref{fig:block}. We will compare the performance of our HOP blocks with global \& local self-attention in the ablation study.

With aforementioned HOP blocks, we build the hierarchical opacity propagation structure for alpha matte estimation as depicted in Figure \ref{fig:diag}(b). A HOP structure involves a global HOP block and multiple local HOP blocks. In the schematic diagram, we omit deconvolution blocks between HOP blocks to display how the opacity information is propagated hierarchically. The bottom global HOP block will perform a global opacity broadcast on  feature maps from the bottleneck, where the feature map contains more semantic but less textural messages. 
It is intuitive to transmit semantic feature globally to leverage the whole information all over the image. Subsequently, local HOP blocks are incorporated into the network between deconvolution stages, where more textural messages are represented in the high-resolution feature map. Thus it is of great motivation to impel local HOP block to only attend to the neighborhood of each query point for textural information extraction. With our HOP structure, the opacity is  propagated across different feature levels, from semantic to textural features and from low-resolution to high-resolution ones.

Moreover, the proposed HOP structure can be considered as a 4-layer graph convolutional network \cite{kipf2016semi} with different  graph in each layer and number of nodes is variable in different stages of the network. The graph in global HOP block is a complete graph and the ones in local HOP blocks are sparse. All edge weights are computed by attention mechanism like graph attention networks \cite{velivckovic2017graph}.
\subsection{Positional Encoding}
\begin{figure}[t]
	\centering	 
	\subfloat[SI-PE]{
		\centering
		\includegraphics[width=.22\linewidth]{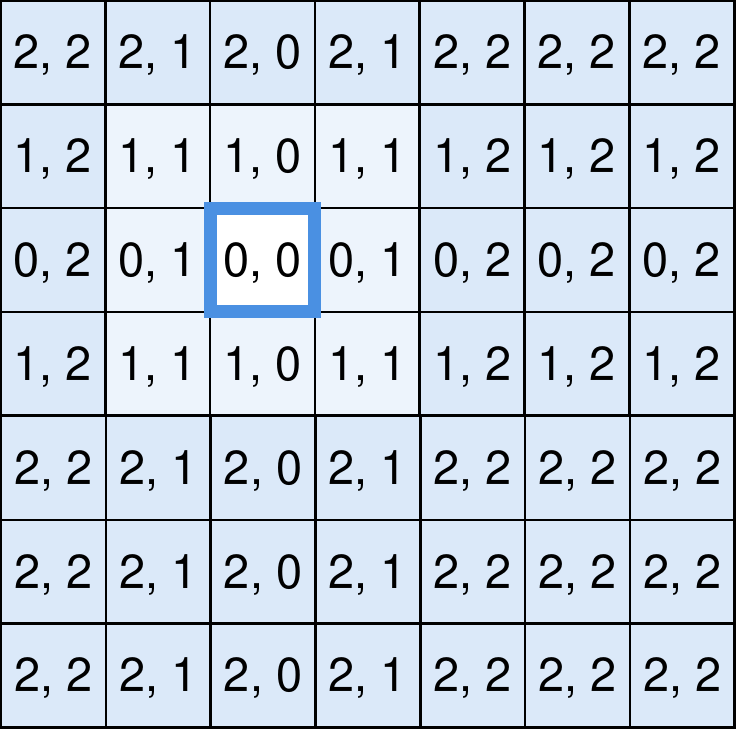}
	}
	\centering	
	\subfloat[R-PE]{
		\centering
		\includegraphics[width=.22\linewidth]{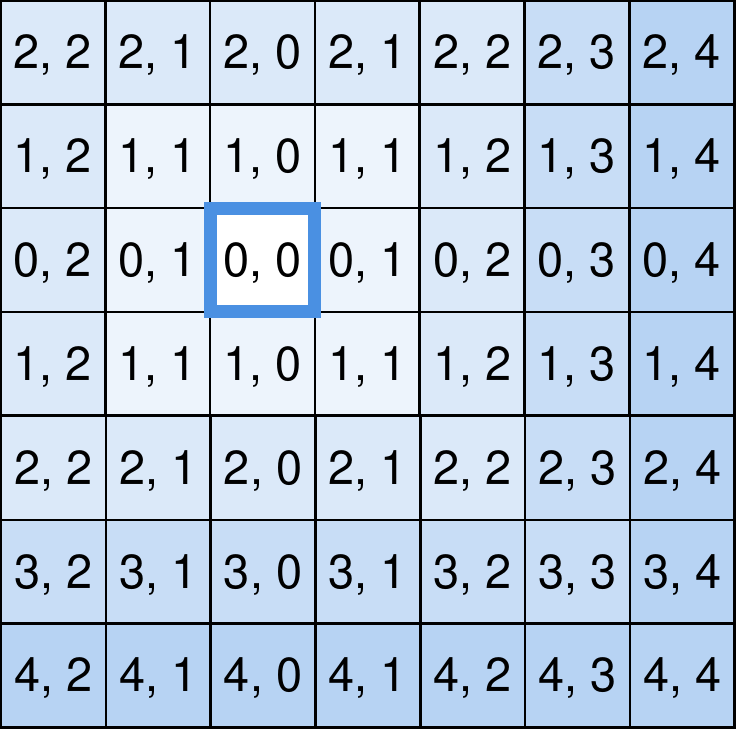}
	}	
	\centering	
	\subfloat[A-PE]{
		\centering
		\includegraphics[width=.22\linewidth]{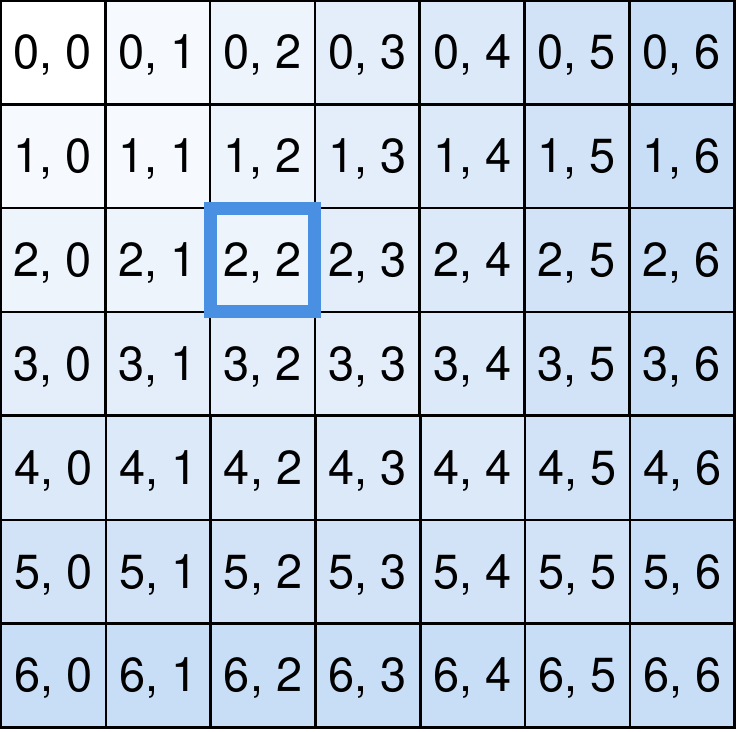}
	}
	\centering	
	\subfloat[LR-PE]{
		\centering
		\includegraphics[width=.22\linewidth]{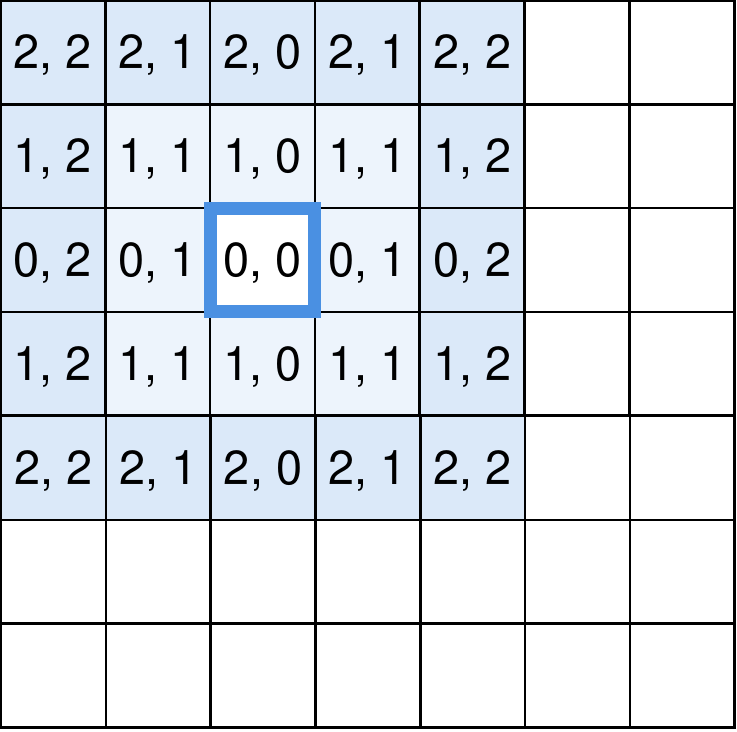}
	}
	\caption{The illustration of different positional encoding methods. The blue rectangle indicates the position of query. The ordered pair in each feature point is (row offset, column offset). SI-PE: scale-insensitive positional encoding, R-PE: relative positional encoding, A-PE: absolute positional encoding, LR-PE: local relative positional encoding.}
	\label{fig:PE}
\end{figure}

Positional encoding always yields gains with self-attention mechanism in several previous work \cite{vaswani2017attention,dai2019transformer,ramachandran2019stand}. We now describe the positional encoding incorporated in our approach. We employ two different positional encoding methods, scale-insensitive position encoding for global HOP block and local relative position encoding for local HOP block. The illustrations of different positional encoding methods are shown in Figure \ref{fig:PE}.

\subsubsection{Scale-insensitive Position Encoding}
Vaswani \textit{et al.} \cite{vaswani2017attention} introduced the positional encoding into transformer to improve the performance in natural language processing tasks. Transformer-xl \cite{dai2019transformer} further extended the absolute positional encoding to relative positional encoding. Intuitively, we can divide the embedding into row and column encodings to extend the relative or absolute positional encoding to a 2-dimensionality encoding for image matting. However, a fatal obstacle of the previous positional encoding is that the neighborhood size of the attention must be fixed. Once the input image size is larger than training ones, there will be some new positional embeddings which never appeared in training. In image matting, it is extremely common that the testing image is larger than the training image patches. To address this issue, we propose a scale-insensitive position encoding for global HOP block. 

In our scale-insensitive positional encoding, we define a radius $s$ of the neighborhood. Any point located beyond the radius shares the same positional encoding, and for the points within radius $s$ we use the relative positional encoding  (see the illustration in Figure \ref{fig:PE}(a)). Then the global HOP block can be written as:
\begin{equation}
\begin{aligned}
e_d =& \begin{cases} W_{PE}\,r_d  \quad d\le s;\\ W_{PE}\,r_s   \quad otherwise,\end{cases} \\
a_{(i,j),(x,y)} =& \mathop{\mathrm{softmax}}_{(x,y)}(\frac{q_{(i,j)}^Tk_{(x,y)}}{\|q_{(i,j)}\|\|k_{(x,y)}\|}\\ & + \frac{q_{(i,j)}^T}{\|q_{(i,j)}\|}(e_{|i-x|}+e_{|j-y|})),\\
g_{(i,j)} =& W_{out}(\sum_{(x,y)}a_{(i,j),(x,y)}f^O_{(x,y)}) + f^O_{(i,j)},
\end{aligned}
\end{equation}
where we employ the sinusoidal encoding $r_d$ following \cite{vaswani2017attention,dai2019transformer} for simplicity, and select $s=7$ in implementation. With the scale-insensitive positional encoding, our HOP matting can handle input images with any shape and size. In addition to the positional embedding, we also design a trimap embedding to learn whether the foreground, background and unknown area should have different weights in attention. Hence the term  $(e_{|i-x|}+e_{|j-y|})$ is modified to $(e_{|i-x|}+e_{|j-y|}+W_{T}t_{(x,y)})$, where $t_{x,y}$ is the data point at position $(x,y)$ from the resized trimap.

\subsubsection{Local Relative Position Encoding}
For the local HOP block, the neighborhood size is always a constant in the network, which makes the scale-insensitive positional encoding not necessary.
We extent the local relative positional encoding proposed in \cite{ramachandran2019stand} to an direct-invariant version without proposing a novel embedding method. The encoding used in local HOP block is depicted in Figure \ref{fig:PE}(d).

In contrast to previous work with positional encoding \cite{vaswani2017attention,dai2019transformer,ramachandran2019stand}, the positional encodings adopted in our image matting method are both direction-invariant, which means the embedding is only related to the absolute distance along row or column between positions of query and key. This trait is motivated by the fact that natural image matting is more a low-level vision problem of less semantics and it should be rotation-invariant. 

\subsection{Loss Function and Implementation Details}

\begin{table}[t]
	\setlength{\tabcolsep}{.5em}
	\caption{Evaluation results on the resized Composition-1k testing set with different testing interpolations. RI is for \textit{random interpolation} augmentation.}
	\centering
	\small
	\begin{tabular}{l|cccc}  
		\toprule
		Methods & Test Interp.& SAD & MSE($10^{-3}$)  & Grad\\
		\midrule
		\multirow{3}{*}{ground-truth}&nearest& 21.66& 6.6& 13.76 \\
		& bilinear&5.8& 0.4&0.49 \\
		& cubic & 1.1& 0.02& 0.02\\
		\midrule
		\multirow{3}{*}{IndexNet Matting \cite{lu2019indices}}&nearest& 62.43& 23.8& 42.91 \\
		& bilinear&46.35& 14.1&24.41 \\
		& cubic & 45.65& 13.1& 25.47\\
		\midrule
		\multirow{3}{*}{HOP-5x5}&nearest&63.79&25.4&39.85\\
		& bilinear & 49.71 & 19.2 & 27.41\\
		& cubic & 38.45& 11.2  &18.87\\
		\midrule
		\multirow{3}{*}{HOP-5x5 + RI}  &nearest& 53.99 &19.5  &40.81\\
		& bilinear &30.34&6.5 &12.87\\
		& cubic & \textbf{29.42} & \textbf{6.3}&\textbf{12.16}\\
		\bottomrule
	\end{tabular}
	\label{tab:RI}
\end{table}

Our network is only trained with the alpha matte reconstruction loss function which is formulated as the averaged  absolute difference between the estimation and ground truth alpha mattes:
\begin{equation}
\mathcal{L} = \frac{1}{|\mathcal{T}_u|}\sum_{i\in \mathcal{T}_u}|\alpha_i - \alpha^{gt}_i|,
\end{equation}
where $\alpha_i$ is the predicted alpha matte at position $i$, $\alpha^{gt}$ is the ground truth alpha matte and $\mathcal{T}_u$ is a set of unknown pixels in the trimap.

We select the first 11 blocks of ResNet-34 \cite{he2016deep} pretrained on ImageNet \cite{russakovsky2015imagenet} as our backbone in the opacity encoder. As for appearance encoder, we opt for a stack of stride convolutional layers to extract more low-level information.
The network is trained on the foreground images from Adobe Image Matting dataset \cite{xu2017deep} and background images from MS COCO dataset \cite{lin2014microsoft}. We follow the basic data augmentation proposed in \cite{li2020natural}. We normalize the training phase by  both batch normalization \cite{ioffe2015batch} and spectral normalization \cite{miyato2018spectral}.  The optimization is performed by Adam optimizer \cite{kingma2014adam}.  
The model is trained in FP16 precision following \cite{he2019bag}. 
Warmup \cite{goyal2017accurate} and cosine
decay \cite{loshchilov2016sgdr}  are applied in our training.

\begin{figure}[t]
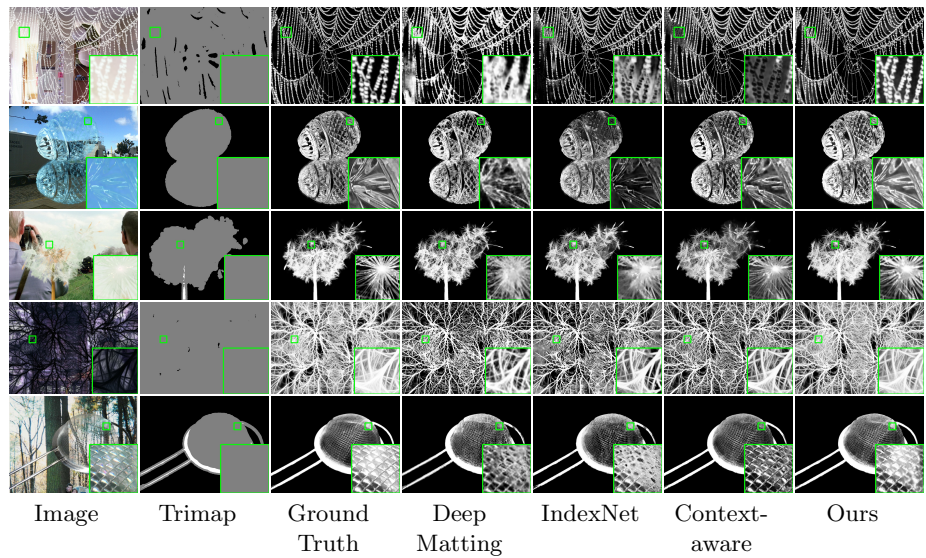

	\small
	\begin{center}	
		\readlist\mylist{cobweb-morgentau-dew-dewdrop-52501_19,crystal-82296_1920_0,dandelion-1335575_1920_7,network-579555_1920_19}
		\readlist\mylistt{Image,Trimap,Ground Truth,Deep Matting,IndexNet,Context-aware,Ours}
		\foreachitem\t\in\mylist{		
			\foreachitem\tt\in\mylistt{
				\centering		 
				\pbox{0.125\textwidth}{
					\centering
					\includegraphics[width=.14\textwidth]{figure/compare/"\tt"/\t.jpg}
				}
			}\\
		}	
		\foreachitem\tt\in\mylistt{
			\centering		  
			\pbox[t]{0.125\textwidth}{
				\includegraphics[width=0.14\textwidth]{figure/compare/"\tt"/sieve-641426_1920_16.jpg} \\ 
				\centering \tt
			}
		}
	\end{center}
	\caption{The visual comparison results on Adobe Composition-1k testing set \cite{xu2017deep}.}
	\label{fig:adobe}
\end{figure}

\begin{table}[t]
\setlength{\tabcolsep}{0.5em}
\caption{The quantitative results on Composition-1k testing set. PosE stands for \textit{positional encoding}, and TriE is for \textit{trimap embedding} in HOP blocks and RI for \textit{random interpolation} augmentation. The variants of our approaches are emphasized in italic. Best results are in boldface.}
\centering
\small
\begin{tabular}{l|cccc}  
	\toprule
	Methods & SAD & MSE($10^{-3}$) & Grad &Conn\\
	\midrule
	DCNN Matting \cite{cho2019deep}& 161.4& 87 &115.1& 161.9 \\
	Learning Based Matting \cite{zheng2009learning}&113.9& 48 &91.6 &122.2\\ 
	Information-flow Matting \cite{aksoy2017designing} & 75.4& 66& 63.0&-\\
	Deep Matting \cite{xu2017deep} &50.4& 14& 31.0& 50.8\\
	IndexNet Matting \cite{lu2019indices} &45.8&	13&	25.9&	43.7\\	
	AdaMatting \cite{cai2019disentangled}&41.7&10  &16.8 &-\\
	SampleNet Matting \cite{samplenet} &40.35&	9.9&	-&	-\\	
	GCA Matting \cite{li2020natural} &35.28&	9.1&	16.9&	32.5\\
	Context-aware Matting \cite{hou2019context}& 35.8 & 8.2& 17.3& 33.2\\
	\midrule
	\textit{HOP-5x5}& 34.82& 9.0 & 16.01& 32.04\\
	
	\textit{HOP-9x9}& {33.44}&{8.2}& {15.62}&{30.44} \\
	\textit{w/o HOP + RI}& 30.45& 6.8 & 13.23& 26.81\\
	\textit{HOP-5x5 + PosE + TriE + RI} & {28.12}& {5.8} & {11.36}& {24.13}\\
	\textit{HOP-9x9 + PosE + TriE + RI}& \textbf{27.80}& \textbf{5.7} & \textbf{11.25}& \textbf{23.73}\\
	\bottomrule
\end{tabular}
\label{tab:adobe}
\end{table}

\begin{table}[t]
\setlength{\tabcolsep}{.15em}
\caption{Our scores on the alphamatting.com benchmark. S, L and U denote three trimap types, small, large and user, included in the benchmark.}
\scriptsize
\centering
\begin{tabular}{l|c|ccc|c|ccc|c|ccc}
	\toprule
	\multirow{2}{*}{Average Rank} & \multicolumn{4}{c|}{SAD}& \multicolumn{4}{c|}{MSE}& \multicolumn{4}{c}{Gradient Error}\\
	& Overall&S&L&U& Overall&S&L&U& Overall&S&L&U\\
	\midrule
	Ours&	\textbf{5.3}&	\textbf{5.8}	&\textbf{4}&	\textbf{6}&	\textbf{7}&	6.9&\textbf{5.4}	&8.6&\textbf{5.4}	&6.4&	\textbf{4.6}&	\textbf{5.1} \\
	
	AdaMatting \cite{cai2019disentangled} &7&	6.1	&6.1&	8.8 &8&	\textbf{5.8}	&7.4&	10.8&7.6&	\textbf{4.5}&	5.3&	13\\
	
	SampleNet Matting \cite{samplenet} &	7.8&5.9&	7.4&	10 &	9.1	&5.9&	9.1	&12.3&	9.1	&5.3&	6.9	&15.1\\
	
	GCA Matting \cite{li2020natural}	&8.5&	9.3	&6.1&	10.3&9.3	&9.3&	8.3	&10.5&7.3&	7.3&	6.1&	8.5	 \\
	
	Deep Matting \cite{xu2017deep}&10.1&	11.4&	9.4	&9.5&13	&11.6&	11.8&	15.6	&17.5&	14.5&	14.1&	24\\
	
	Information-flow matting \cite{aksoy2017designing}&12.2&	13.3&	12.9&	10.4&13.8&	16.3&	13&	12&	20.1&	23	&18.8&	18.6\\
	
	IndexNet Matting \cite{lu2019indices}&13.3&	15.5&	11.9&	12.4	&16.9&	19.4&	15.4&	15.9	&12.5&	11.4&	11&	15.3	\\
	
	AlphaGAN \cite{cai2019disentangled}&14.8&	15.5&	15&	13.8&18&	18.3&	19&	16.6&17.2&	16.1&	15&	20.5\\
	Context-aware Matting \cite{hou2019context}&	17.1&	21&	15&	15.4&11.5&	14.8&	12.8	&\textbf{6.9}&8.7&	9.8&	9.4	&7	\\	
	\bottomrule
	
\end{tabular}
\label{tab:alphamatting}
\end{table}

\begin{table}[t]
\setlength{\tabcolsep}{.5em}
\caption{Parameter numbers and efficiency comparison on Composition-1k testing set on a single NVIDIA RTX 2080 Ti with 11G memory. (Input images of Deep Matting and Context-aware Matting are downsampled with a factor of 0.8.)}
\centering
\small
\begin{tabular}{l|cc}  
	\toprule
	Methods & \# of Parameters & Mean Time \\
	\midrule
	Deep Matting \cite{xu2017deep} &	130.6 M & 0.245 s \\
	Context-aware Matting \cite{hou2019context} & 107.5 M & 3.915 s\\
	IndexNet Matting \cite{lu2019indices} &	\textbf{6.0 M} &\textbf{ 0.182 s}\\	
	\midrule
	\textit{HOP-1x1} & 25.3 M & \textbf{0.182 s} \\
	\textit{HOP-5x5} & 25.3 M & 0.255 s \\
	\textit{HOP-9x9} & 25.3 M & 0.339 s \\
	\bottomrule
\end{tabular}
\label{tab:eff}
\end{table}
\section{Experiments}
We conduct extensive experiments and ablation study to demonstrate the effectiveness of our proposed HOP matting. We report the empirical results of our HOP matting on two widely used datasets, the Composition-1k testing set \cite{xu2017deep} and alphamatting.com dataset \cite{rhemann2009perceptually}. The results are evaluated under mean squared error (MSE), sum of absolute difference (SAD), gradient error (Grad) and connectivity error (Conn) as \cite{rhemann2009perceptually} suggested. We also visualize the attention in our HOP structure for a better understanding.

\subsection{Random Interpolation Augmentation}

Empirically, the performance of deep image matting methods is sensitive to the image resize. It is because that typical natural image matting approaches attend to the detailed texture information in images, and the resize operation may blur edges or high frequency information and lead to a deterioration in the performance. Therefore, most of the matting methods are evaluated on the original image without any resize operation. In Context-aware Matting \cite{hou2019context}, the authors claimed that different image formats of foreground and background will introduce subtle artifacts into composited training images, which can help the network to distinguish foreground from background.  Analogously, in this section, we will show some new observations that deep neural network based matting methods are sensitive to the interpolation algorithm and introduce the random interpolation augmentation used in our method. 

We conduct an empirical experiment to support this observation on the Composition-1k testing set \cite{xu2017deep}. Firstly, we upsample RGB images with a factor of 1.5 by a selected interpolation algorithm and then downsample images to their original size by the same interpolation algorithm. More concretely, supposing that the RGB image is 800 x 800 and the selected testing interpolation is bilinear, we resize the image to 1200 x 1200 by bilinear interpolation and resize the new image back to 800 x 800 by bilinear interpolation afterwards. Finally, the resized RGB image is feed forward to the network and we evaluate the error between prediction and original ground truth. It is worth noting that we do not begin with a downsampling followed by an upsampling operation because downsampling first will lead to more information loss.

Evaluation results are reported in Table \ref{tab:RI}. We provide the method \textit{ground-truth} as a reference. Method \textit{ground-truth} means that we directly resize the ground truth alpha matte image without any estimation and then  calculate the error between resized and unresized ground truth images. These results reveal the error introduced by interpolation itself, which can be seen as an ideal lower bound of these evaluation results. \textit{HOP-5x5} stands for our baseline HOP model with 5x5 neighborhood in the local HOP block and without positional encoding or trimap embedding in HOP blocks. From Table \ref{tab:RI}, we can notice that most of the result gaps between different testing interpolations are larger than the \textit{ground-truth} lower bound. In other words, different interpolation algorithms can introduce more error in inference than the interpolation itself. We can also see the gap between bilinear and cubic of our \textit{HOP-5x5} method is larger than IndexNet Matting \cite{lu2019indices}. Our explanation is that, in the data augmentation of training set, we resize the background image to the same size as foreground by cubic interpolation following Deep Matting \cite{xu2017deep}. This fixed interpolation augmentation makes our model fit cubic interpolation much better than the others. 

Based on empirical observation aforementioned, we introduce the random interpolation augmentation into our method. In the data preprocessing of training phase, we randomly select a interpolation algorithm with equal probability for any resize operation. Therefore, the composited images in a training mini-batch may be generated by different interpolations. Furthermore, the foreground, background and alpha matte images can be resize with different algorithm before the composition. As Table \ref{tab:RI} shows, the training with random interpolation not only improves the performance but also mitigates the error gap between bilinear and cubic interpolations. 

\subsection{Results on Composition-1k Testing Set}
The Composition-1k testing set \cite{xu2017deep} contains 1000 composed images from 50 distinct foregrounds. We compare our method with the other top-tier natural image matting approaches quantitatively as the results shown in Table \ref{tab:adobe}.  Variant \textit{w/o HOP + RI} indicates the backbone network without any HOP block and trained with random interpolation augmentation. All our variants outperform state-of-the-art methods. Some qualitative results on Composition-1k testing set are displayed in Figure \ref{fig:adobe}. The results of Deep Matting \cite{xu2017deep} are generated from the source code and pretrained model provided by IndexNet Matting \cite{lu2019indices}.

Furthermore, we compare the number of parameters and the model efficiency with some of the state-of-the-art methods in Table \ref{tab:eff}. We evaluate the mean inference time of each image in Composition-1k testing set on a single NVIDIA RTX 2080 Ti GPU. Notably, Context-aware Matting \cite{hou2019context} and Deep Matting \cite{xu2017deep} require more than 11G GPU memory to estimate the alpha mattes of high-resolution images from Composition-1k testing set. Thus we downsample the input image with a factor of 0.8 for these two methods.

\begin{table}[t]
	\setlength{\tabcolsep}{0.5em}
	\caption{Ablation study on different HOP blocks on the Composition-1k testing set. HOP-Local-$k$ indicates the $k$-th local HOP block in the decoder.}
	\centering
	\small
	\begin{tabular}{cccc|ccc}  
		\toprule
		HOP-Global &HOP-Local-1 &HOP-Local-2 &HOP-Local-3& SAD& MSE($10^{-3}$) \\
		\midrule
		&&& & 37.89  & 10.05   \\
		\checkmark&&& & 36.96 & 9.89  \\
		\checkmark&\checkmark&&& 37.36  & 9.86 \\
		\checkmark&\checkmark&\checkmark& & 36.11 & 9.32 \\
		\checkmark&\checkmark&\checkmark&\checkmark& \textbf{34.82} & \textbf{8.99}\\
		\midrule
		\multicolumn{4}{c|}{Global \& Local Self-attention}&  35.97 & 9.24 \\
		\bottomrule
	\end{tabular}
	\label{tab:block}
\end{table}

\begin{table}[t]	
	\setlength{\tabcolsep}{0.5em}
	\caption{Ablation study on positional encoding and trimap embedding and random interpolation augmentation on the Composition-1k testing set.}
	\centering
	\small
	\begin{tabular}{l|ccc|cccc}  
		\toprule
		Method & PosE &TriE&RI& SAD& MSE($10^{-3}$) &Grad& Conn\\
		\midrule
		\multirow{5}{*}{HOP-5x5}&&&& 34.82& 9.0 & 16.01& 32.04\\
		&\checkmark&& & 33.87& 8.2 & 15.34& 31.09\\
		&\checkmark&\checkmark&&  34.15 & 8.1 & 15.33 & 31.45\\
		&&&\checkmark & {29.01}& {6.2} & {11.93}& {25.09}\\
		&\checkmark&\checkmark&\checkmark & \textbf{28.12}& \textbf{5.8} & \textbf{11.36}& \textbf{24.13}\\
		\bottomrule
	\end{tabular}
	\label{tab:emb}
\end{table}
\subsection{Results on Alphamatting.com Dataset}
The alphamatting.com dataset contains eight test images for online benchmark evaluation. Each test image has three different trimaps (\textit{i.e.} "small", "large" and "user").  We provide the average rank value of our proposed approach on alphamatting.com benchmark in Table \ref{tab:alphamatting}. The \textit{Overall} rank is an average rank over all three trimap types for each evaluation metric. As the ranking shows in Table \ref{tab:alphamatting}, our HOP matting outperform the other state-of-the-art approaches under different evaluation metrics.

%
\subsection{Ablation Study}
To validate the efficacy of each component in HOP matting, we conduct three different experiments on the Composition-1k testing set.
We first evaluate the \textit{HOP-5x5} model by removing different HOP blocks. From the results reported in Table \ref{tab:block}, we can notice that the hierarchical opacity propagation structure is capable of improving the performance of networks in image matting. \textit{Global} \& \textit{Local Self-attention} indicates the method that replaces global HOP block with global self-attention and replaces local HOP block with local self-attention. 
In the second ablation study, we reveal the effect of introducing positional encoding, trimap embedding and random interpolation augmentation into our method. The quantitative results evaluated on Composition-1k testing set \cite{xu2017deep} is provided in Table \ref{tab:emb}. We also provide the evaluation of different neighborhood window size in supplementary material.


\subsection{Visualization of HOP Structure}
\begin{figure}[t]
	\centering	 	
	\pbox[t]{.23\textwidth}{
		\includegraphics[width=.23\textwidth]{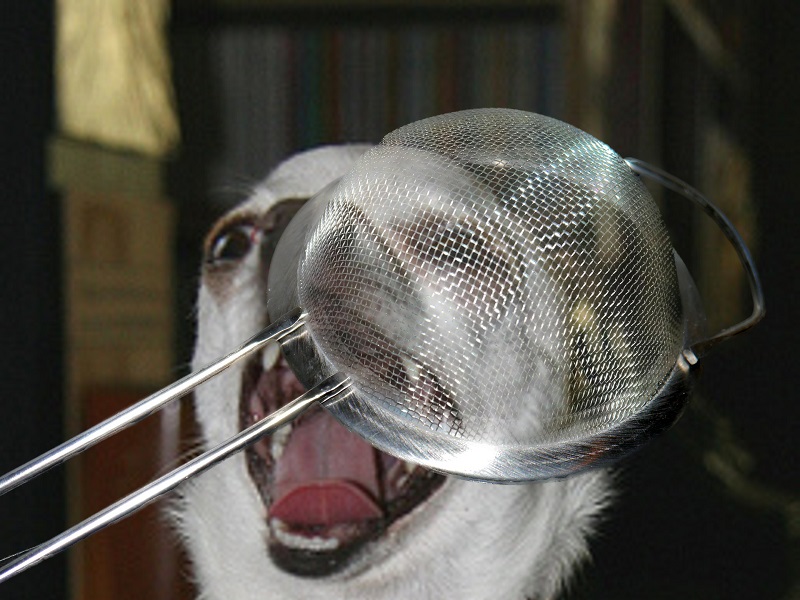} \\
		\centering Image
	}	 
	\pbox[t]{.23\textwidth}{
		\includegraphics[width=.23\textwidth]{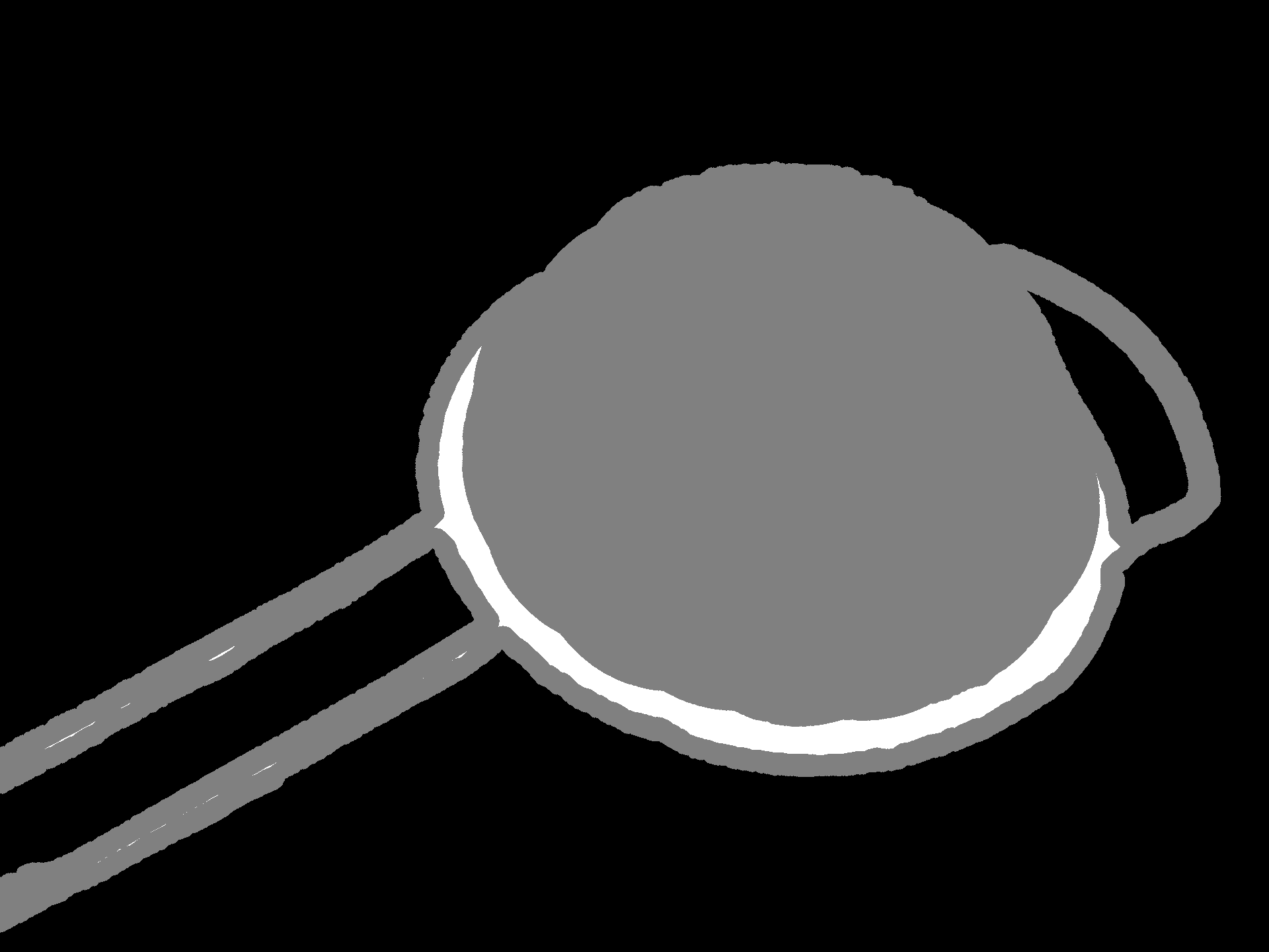}\\
		\centering Trimap
	}	 
	\pbox[t]{.23\textwidth}{
		\includegraphics[width=.23\textwidth]{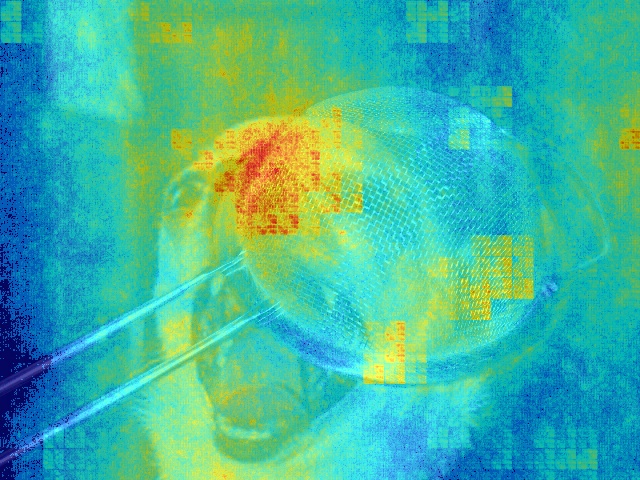} \\
		\centering Ours
	}
	\pbox[t]{.23\textwidth}{
		\includegraphics[width=.23\textwidth]{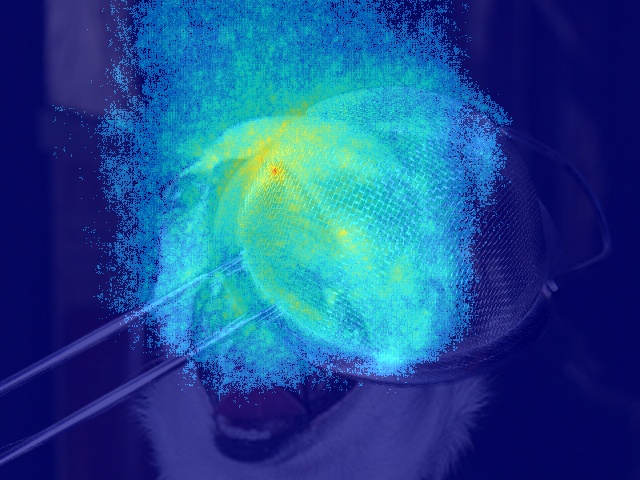}\\
		\centering Ours without  HOP Block
	}
	
	\caption{The visualization of attention in our HOP structure on input images.}
	\label{fig:vis}
\end{figure}

Visualizing the attention of HOP structure is a convenient way to understand how the opacity information is propagated hierarchically in our approach. To this end, we visualize where our model attend to by the gradient map on the input image. 
We randomly select a pixel in the unknown region from the alpha matte prediction. Then a large loss is assigned to this single pixel and all the other pixels of the prediction are assumed to be perfectly correct without any loss. Afterwards, the back-propagation is executed and we propagate the gradient backward to the input image. The gradient map reveals how each pixel of the input image is related to the selected alpha matte pixel in the prediction. We show the gradient map of an image from Composition-1k testing set \cite{xu2017deep} in Figure \ref{fig:vis}. The results without HOP block are generated from the model we train for ablation study reported in Table \ref{tab:block}.
As we can observe from Figure \ref{fig:vis}, the model with HOP block is able to aggregate information all over the input image and pay more attention to the area with similar appearance, while the model without HOP blocks, conversely, focuses more on a local region around the selected prediction point.

\section{Conclusions and Future Work}
In this paper, we proposed a HOP matting network for image matting. Our method utilizes local and global HOP blocks to achieve the hierarchical opacity propagation across feature maps at different semantic levels. The experimental results demonstrate the superiority of the proposed HOP matting. Furthermore, the effectiveness of our positional encoding and random interpolation augmentation are verified by the ablation study.
Considering the success of fully self-attention networks
 \cite{ramachandran2019stand}, it is a promising future work to investigate how the fully HOP block 
networks works in
 matting.
  Another interesting future work is a hybrid network of stacked local HOP blocks and  fully convolutional blocks.   

\clearpage
%
%
\bibliographystyle{splncs04}
\bibliography{egbib}

\begin{thebibliography}{10}
\providecommand{\url}[1]{\texttt{#1}}
\providecommand{\urlprefix}{URL }
\providecommand{\doi}[1]{https://doi.org/#1}

\bibitem{aksoy2017designing}
Aksoy, Y., Ozan~Aydin, T., Pollefeys, M.: Designing effective inter-pixel
  information flow for natural image matting. In: CVPR (2017)

\bibitem{bahdanau2014neural}
Bahdanau, D., Cho, K., Bengio, Y.: Neural machine translation by jointly
  learning to align and translate. arXiv preprint arXiv:1409.0473  (2014)

\bibitem{cai2019disentangled}
Cai, S., Zhang, X., Fan, H., Huang, H., Liu, J., Liu, J., Liu, J., Wang, J.,
  Sun, J.: Disentangled image matting. In: Proceedings of the IEEE
  International Conference on Computer Vision. pp. 8819--8828 (2019)

\bibitem{chen2013knn}
Chen, Q., Li, D., Tang, C.K.: Knn matting. IEEE TPAMI  (2013)

\bibitem{chen2018semantic}
Chen, Q., Ge, T., Xu, Y., Zhang, Z., Yang, X., Gai, K.: Semantic human matting.
  In: ACM MM (2018)

\bibitem{child2019generating}
Child, R., Gray, S., Radford, A., Sutskever, I.: Generating long sequences with
  sparse transformers. arXiv preprint arXiv:1904.10509  (2019)

\bibitem{cho2019deep}
Cho, D., Tai, Y.W., Kweon, I.S.: Deep convolutional neural network for natural
  image matting using initial alpha mattes. IEEE TIP  \textbf{28}(3),
  1054--1067 (2019)

\bibitem{chuang2001bayesian}
Chuang, Y.Y., Curless, B., Salesin, D.H., Szeliski, R.: A bayesian approach to
  digital matting. In: CVPR (2). pp. 264--271 (2001)

\bibitem{dai2019transformer}
Dai, Z., Yang, Z., Yang, Y., Cohen, W.W., Carbonell, J., Le, Q.V.,
  Salakhutdinov, R.: Transformer-xl: Attentive language models beyond a
  fixed-length context. arXiv preprint arXiv:1901.02860  (2019)

\bibitem{feng2016cluster}
Feng, X., Liang, X., Zhang, Z.: A cluster sampling method for image matting via
  sparse coding. In: European Conference on Computer Vision. pp. 204--219.
  Springer (2016)

\bibitem{gastal2010shared}
Gastal, E.S., Oliveira, M.M.: Shared sampling for real-time alpha matting. In:
  Computer Graphics Forum. vol.~29, pp. 575--584. Wiley Online Library (2010)

\bibitem{goyal2017accurate}
Goyal, P., Doll{\'a}r, P., Girshick, R., Noordhuis, P., Wesolowski, L., Kyrola,
  A., Tulloch, A., Jia, Y., He, K.: Accurate, large minibatch sgd: Training
  imagenet in 1 hour. arXiv preprint arXiv:1706.02677  (2017)

\bibitem{he2011global}
He, K., Rhemann, C., Rother, C., Tang, X., Sun, J.: A global sampling method
  for alpha matting. In: CVPR 2011. pp. 2049--2056. IEEE (2011)

\bibitem{he2010fast}
He, K., Sun, J., Tang, X.: Fast matting using large kernel matting laplacian
  matrices. In: 2010 IEEE Computer Society Conference on Computer Vision and
  Pattern Recognition. pp. 2165--2172. IEEE (2010)

\bibitem{he2016deep}
He, K., Zhang, X., Ren, S., Sun, J.: Deep residual learning for image
  recognition. In: CVPR (2016)

\bibitem{he2019bag}
He, T., Zhang, Z., Zhang, H., Zhang, Z., Xie, J., Li, M.: Bag of tricks for
  image classification with convolutional neural networks. In: CVPR (2019)

\bibitem{hochreiter1997long}
Hochreiter, S., Schmidhuber, J.: Long short-term memory. Neural computation
  \textbf{9}(8),  1735--1780 (1997)

\bibitem{hou2019context}
Hou, Q., Liu, F.: Context-aware image matting for simultaneous foreground and
  alpha estimation. In: Proceedings of the IEEE International Conference on
  Computer Vision. pp. 4130--4139 (2019)

\bibitem{huang2019interlaced}
Huang, L., Yuan, Y., Guo, J., Zhang, C., Chen, X., Wang, J.: Interlaced sparse
  self-attention for semantic segmentation. arXiv preprint arXiv:1907.12273
  (2019)

\bibitem{ioffe2015batch}
Ioffe, S., Szegedy, C.: Batch normalization: Accelerating deep network training
  by reducing internal covariate shift. arXiv preprint arXiv:1502.03167  (2015)

\bibitem{kingma2014adam}
Kingma, D.P., Ba, J.: Adam: A method for stochastic optimization. arXiv
  preprint arXiv:1412.6980  (2014)

\bibitem{kipf2016semi}
Kipf, T.N., Welling, M.: Semi-supervised classification with graph
  convolutional networks. arXiv preprint arXiv:1609.02907  (2016)

\bibitem{lee2011nonlocal}
Lee, P., Wu, Y.: Nonlocal matting. In: CVPR 2011. pp. 2193--2200. IEEE (2011)

\bibitem{levin2008closed}
Levin, A., Lischinski, D., Weiss, Y.: A closed-form solution to natural image
  matting. IEEE TPAMI  (2008)

\bibitem{li2020natural}
Li, Y., Lu, H.: Natural image matting via guided contextual attention. arXiv
  preprint arXiv:2001.04069  (2020)

\bibitem{lin2014microsoft}
Lin, T.Y., Maire, M., Belongie, S., Hays, J., Perona, P., Ramanan, D.,
  Doll{\'a}r, P., Zitnick, C.L.: Microsoft coco: Common objects in context. In:
  ECCV. pp. 740--755. Springer (2014)

\bibitem{lin2017structured}
Lin, Z., Feng, M., Santos, C.N.d., Yu, M., Xiang, B., Zhou, B., Bengio, Y.: A
  structured self-attentive sentence embedding. arXiv preprint arXiv:1703.03130
   (2017)

\bibitem{loshchilov2016sgdr}
Loshchilov, I., Hutter, F.: Sgdr: Stochastic gradient descent with warm
  restarts. arXiv preprint arXiv:1608.03983  (2016)

\bibitem{lu2019indices}
Lu, H., Dai, Y., Shen, C., Xu, S.: Indices matter: Learning to index for deep
  image matting. In: ICCV (2019)

\bibitem{lutz2018alphagan}
Lutz, S., Amplianitis, K., Smolic, A.: Alphagan: Generative adversarial
  networks for natural image matting. In: BMVC (2018)

\bibitem{miyato2018spectral}
Miyato, T., Kataoka, T., Koyama, M., Yoshida, Y.: Spectral normalization for
  generative adversarial networks. arXiv preprint arXiv:1802.05957  (2018)

\bibitem{ramachandran2019stand}
Ramachandran, P., Parmar, N., Vaswani, A., Bello, I., Levskaya, A., Shlens, J.:
  Stand-alone self-attention in vision models. arXiv preprint arXiv:1906.05909
  (2019)

\bibitem{rhemann2009perceptually}
Rhemann, C., Rother, C., Wang, J., Gelautz, M., Kohli, P., Rott, P.: A
  perceptually motivated online benchmark for image matting. In: CVPR (2009)

\bibitem{russakovsky2015imagenet}
Russakovsky, O., Deng, J., Su, H., Krause, J., Satheesh, S., Ma, S., Huang, Z.,
  Karpathy, A., Khosla, A., Bernstein, M., et~al.: Imagenet large scale visual
  recognition challenge. IJCV  \textbf{115}(3),  211--252 (2015)

\bibitem{shen2016deep}
Shen, X., Tao, X., Gao, H., Zhou, C., Jia, J.: Deep automatic portrait matting.
  In: ECCV (2016)

\bibitem{tang2019very}
Tang, H., Huang, Y., Fan, Y., Zeng, X., et~al.: Very deep residual network for
  image matting. In: 2019 IEEE International Conference on Image Processing
  (ICIP). pp. 4255--4259. IEEE (2019)

\bibitem{samplenet}
Tang, J., Aksoy, Y., \"Oztireli, C., Gross, M., Ayd{\i}n, T.O.: Learning-based
  sampling for natural image matting. In: Proc. CVPR (2019)

\bibitem{vaswani2017attention}
Vaswani, A., Shazeer, N., Parmar, N., Uszkoreit, J., Jones, L., Gomez, A.N.,
  Kaiser, {\L}., Polosukhin, I.: Attention is all you need. In: NIPS (2017)

\bibitem{velivckovic2017graph}
Veli{\v{c}}kovi{\'c}, P., Cucurull, G., Casanova, A., Romero, A., Lio, P.,
  Bengio, Y.: Graph attention networks. arXiv preprint arXiv:1710.10903  (2017)

\bibitem{wang2007optimized}
Wang, J., Cohen, M.F.: Optimized color sampling for robust matting. In: 2007
  IEEE Conference on Computer Vision and Pattern Recognition. pp.~1--8. IEEE
  (2007)

\bibitem{wang2008image}
Wang, J., Cohen, M.F., et~al.: Image and video matting: a survey. Foundations
  and Trends{\textregistered} in Computer Graphics and Vision  \textbf{3}(2),
  97--175 (2008)

\bibitem{wang2018non}
Wang, X., Girshick, R., Gupta, A., He, K.: Non-local neural networks. In:
  Proceedings of the IEEE Conference on Computer Vision and Pattern
  Recognition. pp. 7794--7803 (2018)

\bibitem{xingjian2015convolutional}
Xingjian, S., Chen, Z., Wang, H., Yeung, D.Y., Wong, W.K., Woo, W.c.:
  Convolutional lstm network: A machine learning approach for precipitation
  nowcasting. In: Advances in neural information processing systems. pp.
  802--810 (2015)

\bibitem{xu2015show}
Xu, K., Ba, J., Kiros, R., Cho, K., Courville, A., Salakhudinov, R., Zemel, R.,
  Bengio, Y.: Show, attend and tell: Neural image caption generation with
  visual attention. In: International conference on machine learning. pp.
  2048--2057 (2015)

\bibitem{xu2017deep}
Xu, N., Price, B., Cohen, S., Huang, T.: Deep image matting. In: CVPR (2017)

\bibitem{yu2018generative}
Yu, J., Lin, Z., Yang, J., Shen, X., Lu, X., Huang, T.S.: Generative image
  inpainting with contextual attention. In: CVPR (2018)

\bibitem{Zhang2019ALF}
Zhang, Y., Gong, L., Fan, L., Ren, P., Huang, Q., Bao, H., Xu, W.: A late
  fusion cnn for digital matting. In: CVPR (2019)

\bibitem{zheng2009learning}
Zheng, Y., Kambhamettu, C.: Learning based digital matting. In: ICCV (2009)

\bibitem{zhu2019asymmetric}
Zhu, Z., Xu, M., Bai, S., Huang, T., Bai, X.: Asymmetric non-local neural
  networks for semantic segmentation. In: Proceedings of the IEEE International
  Conference on Computer Vision. pp. 593--602 (2019)

\end{thebibliography}
\end{document}